**Title: Variational Autoencoding the Lagrangian Trajectories of Particles in a Combustion System**

**Authors:**

Pai Liu[1, *], Jingwei Gan[2] and Rajan K. Chakrabarty[1, 3]

**Affiliations:**

[1] Center for Aerosol Science and Engineering, Department of Energy, Environmental and Chemical Engineering, Washington University in St. Louis, St. Louis, Missouri 63130, USA

[2] Department of Chemical and Biological Engineering, Illinois Institute of Technology, Chicago, Illinois 60616, USA

[3] McDonnell Center for the Space Sciences, Washington University in St. Louis, St. Louis, Missouri 63130, USA

[*] Address correspondence to: p.liu@wustl.edu

*Version 2*

*Updated on 11 Dec. 2018.*

# ABSTRACT

We introduce a deep learning method to simulate the motion of particles trapped in a chaotic recirculating flame. The Lagrangian trajectories of particles, captured using a high-speed camera and subsequently reconstructed in 3-dimensional space, were used to train a variational autoencoder (VAE) which comprises multiple layers of convolutional neural networks. We show that the trajectories, which are statistically representative of those determined in experiments, can be generated using the VAE network. The performance of our model is evaluated with respect to the accuracy and generalization of the outputs.

# I. INTRODUCTION

The Lagrangian description of motion allows an observer to follow individual particles as they move through space and time [1]. Such a description is commonly adopted when emphasis is placed on the movement of individual particles, specifically the history of their exact locations [2-4]. For example, when modelling colloidal aggregation and surface deposition, millions of numerically simulated particles are allowed to move around according to preset rules, such as Brownian random walk or ballistic motion [5-8]. The model algorithm tracks the Lagrangian motion of these particles and determines the collisions among them when their trajectories overlap [5-8]. Trajectory modelling is not limited to Brownian particle systems. Such technique has been applied to mimicking trajectories for a wide variety of moving objects, such as animals foraging [9-11], human beings using transportation [12,13], and buoys drifting along coastline [14, 15]. The simulated trajectories are subsequently analyzed, and statistical inferences can be made about behaviors in various domains, including search strategy, urban planning, traffic control, and maritime rescue [9-15]. The conventional approaches employed in simulating motion are based on first principle methods, in which the tracer faithfully follows well-defined motion equations [5-8, 13-15]. When the application of a first principle-based model is too difficult or impossible, statistical approaches are commonly adopted as an alternative. For example, correlated random walk models operate with a stochastic input which follows the probability distributions determined in experiments [9, 10, 12, 16].

Recent advancements in deep learning have enabled system simulation via a data driven scheme [17]. The latest studies in this area have showed that deep generative networks succeed in modelling the complex dynamics involved in many physical, chemical, and biophysical systems [18-21]. In this work, we explore the possibility of simulating the stochastic motion of particles using a variational autoencoder (VAE). In such model, deep neural networks are deployed to extract features from the ground truth, and randomly generate new trajectories which share the characteristics of the trajectories determined in experiments [22, 23]. The VAE network comprises a pair of encoder and decoder networks [18, 22, 23] that respectively serve the purposes of converting original input – via dimension reduction – to latent representations and reconstructing the latent variables back to the original form. Stochasticity is introduced in the latent space, commonly as a noise input following a unit Gaussian distribution, so that the generated trajectories manifest diversity [18, 22, 23].

In the following, we first describe the experimental acquisition of the Lagrangian trajectories of particles that are trapped in a flame system introduced in Refs [24-26]. This description is immediately followed by a section detailing the architecture of the VAE model, the design of objective functions, and the optimization procedure. Next, we present the simulation results and evaluate the model performance with respect to two criteria – accuracy and generalization. We conclude this paper by discussing the advantages and limitations of using VAE-based models to simulate trajectories.

## II. METHODS

### A. Experimental acquisition of particle trajectories

Tracking experiments were conducted on soot particles produced using a buoyancy-opposed flame aerosol reactor (BoFAR). Details about the design and schematics of the BoFAR can be found in previous publications [24-26]. In this work, we operate the BoFAR with ethylene and oxygen in a down-fired, non-premixed configuration, which facilitates rapid soot production, aggregation, and gelation [24]. The buoyancy force, opposing the inflow direction, triggers recirculation in the flame body, wherein the soot particles are trapped for a period of several minutes [24-26]. Visually, these particles grow to millimeter size, when they gravitationally settle out.

FIG. 1 shows a schematic diagram of the particle tracking apparatus, along with the setup of a 3-dimenisonal (3-d) Cartesian coordinate ($x$-$y$-$z$) system in the experimental space. A glass mirror (McMaster Carr-2925245C) was placed next to the flame at a 45° angle relative to the $x$-$z$ plane, and a high-speed camera (CASIO EX-FH20) was installed in front of the flame, with a lens pointing opposite to the positive $y$ direction. Such an arrangement facilitated the simultaneous acquisition of the particle trajectories, which were projected onto the orthogonal planes ($x$-$z$ and $y$-$z$), and captured using a single camera. Subsequently, the particle trajectories in 3-d space were readily reconstructed. The camera was operated at 210 frame per seconds yields a unit timescale $t_0 \approx 4.76$ms.

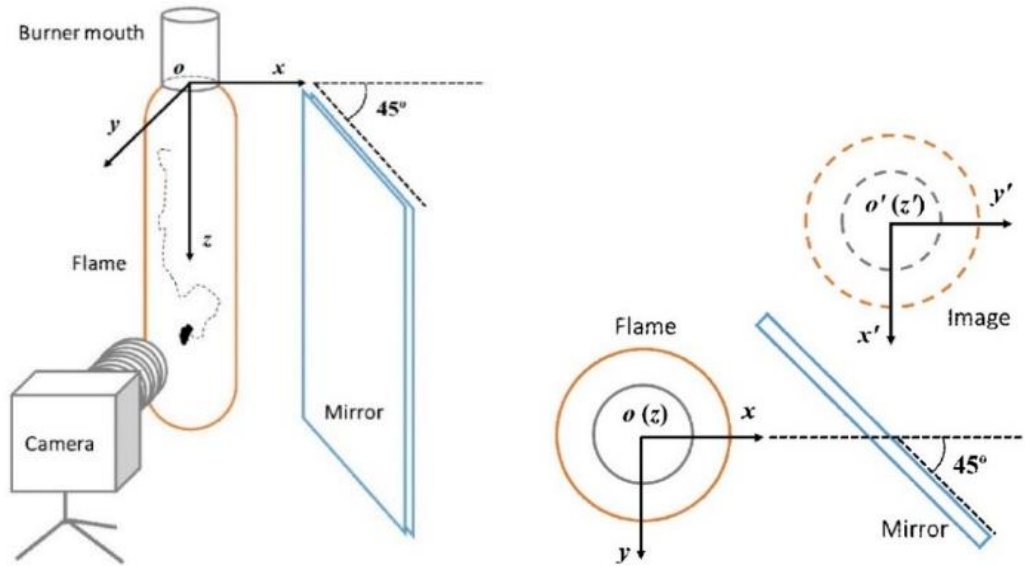

**FIG. 1. Schematic diagram of the particle tracking apparatus in a 3-d Cartesian coordinate system. A top-down view of this arrangement is presented on the right-hand side.**

The target particles were identified in the camera images using a custom-made algorithm in MATLAB. FIG. 2 shows the working of the tracking algorithm. The identification is a simple supervised process in which the algorithm determines whether a pixel belongs to a target particle according to the pixel's brightness value. Subsequently the algorithm calculates the geometric center position (hereafter written as $s$ in a generic manner) of the target particle using the coordinates of all its constituent pixels. When operating the tracking algorithm, we need to manually identify the initial position for each target particle – the $s_1$ at $t_1$. Subsequently for any timestep $(t_i)$ greater than $t_1$, the algorithm opens a scan window near the $s_{i-1}$ and repeat the aforementioned scanning procedure to determine $s_i$.

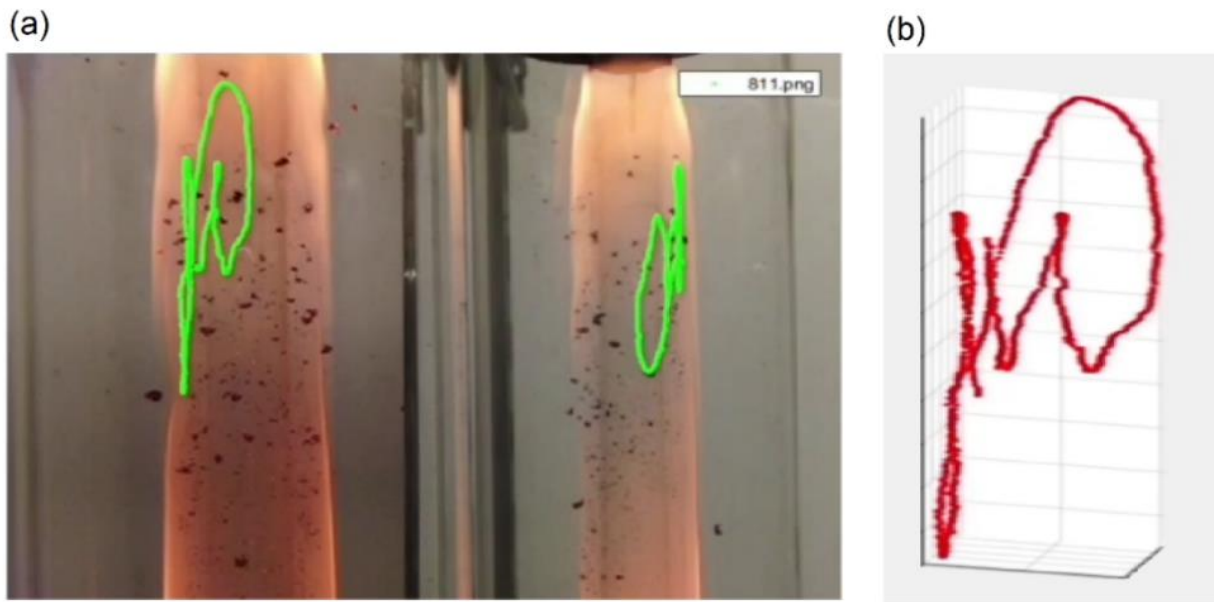

**FIG. 2. Working of the particle tracking algorithm. (a) The image taken by the high-speed camera under the arrangement shown in FIG. 1. The images of the flame projected onto the *x-z* and *y-z* planes are respectively seen on the left and right flanks of the photograph. The trajectory of one target particle is colored in green. (b) A 3-d reconstruction of the trajectory in (a).**

In this work, we captured the 3-d trajectory datasets for fifteen individual particles ($m = 15$), and each tracking lasted for 1100 timesteps, giving a total duration $t_n \approx 5.24s$. The trajectory datasets for these particles were next converted to cylindrical space, $s(r, \theta, z, t)$, which best describes the geometry of our system. In the following, we occasionally use $s(t)$ to denote the cylindrical components $r(t), \theta(t),$ and $z(t)$ in a generic manner, for succinct presentation.

## B. Variational autoencoder architecture

FIG. 3 shows the architecture of the VAE, which has three main components: the encoder network, which comprises three layers of convolutional neural network (CNN); a coding component, which operates in the latent space; and a decoder network, which comprises three layers of transposed CNN (t-CNN). In the following, we describe the detailed structures and functions for each component according to the data flow direction.

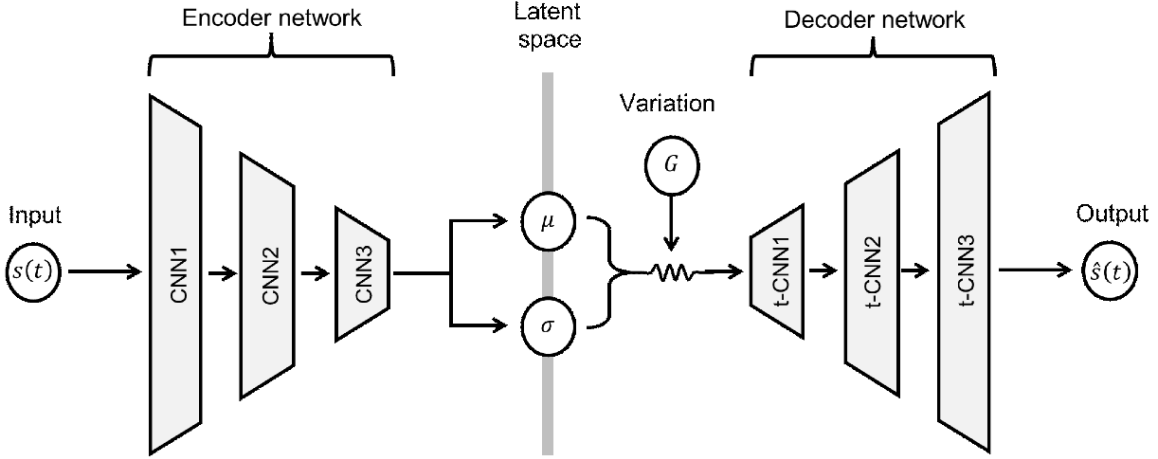

**FIG. 3. Schematic of the VAE architecture. The three main components of the VAE are an encoder network with three CNN layers, a fully-connected neural network in the latent spaces, and a decoder network with three t-CNN layers.**

The experimentally obtained trajectory datasets $s(r,\theta,z,t)$, which serve as inputs of the encoder network, were fed to CNN1. FIG. 4 describes the structure of each encoding CNN layer, along with its treatment of one batch of input dataset as an example. Each layer of encoding CNN extracts features and, at the same time, reduces the dimensions of the data matrices [17,27]. Specifically, CNN1 processes the input $s(r,\theta,z,t)$ dataset using a series of weight matrices, each of which corresponds to a particular characteristic, shown as the colored arrows in FIG. 4. Such characteristic extraction leads to the generation of a lower dimension feature map (FM1), which is next fed to CNN2, and so forth. CNN2 and 3 function in a similar manner, and the output of encoder network is FM3, which is to be used to calculate coding variables in the latent space.

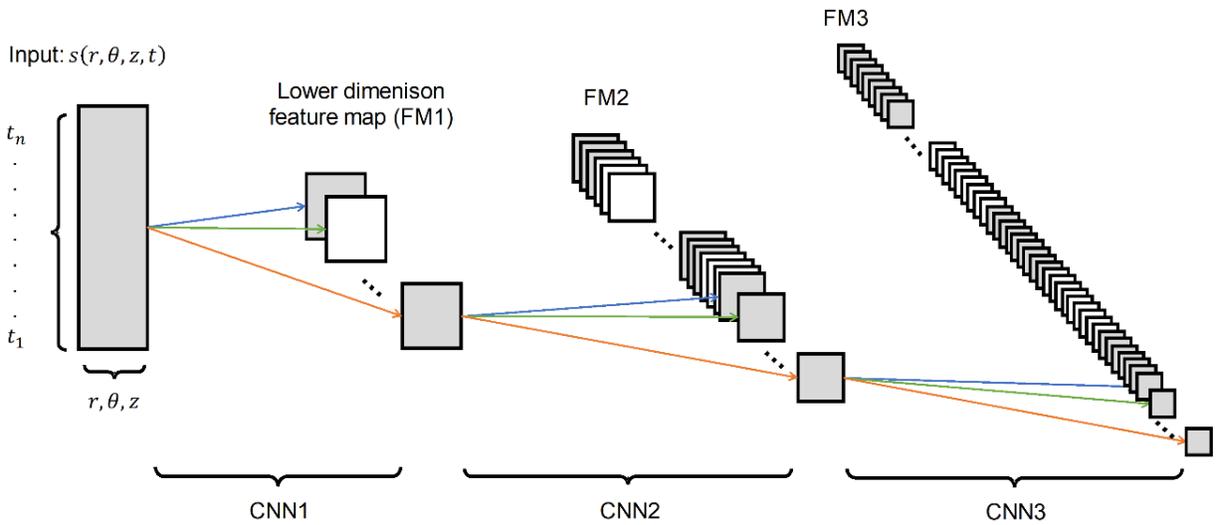

**FIG. 4. Schematic diagram illustrating the structure of the encoding CNN. The rectangle on the far left represents one batch of the experimental dataset, and the following squares represent lower dimension feature maps (FM) that are generated via characteristic extraction. The arrows represent weight matrices used to extract characteristics from the original input (or the FM generated in the previous layer.)**

The output of the encoder network is next fed to the coding component, which comprises two separate fully-connected networks. These two networks individually calculate the mean ($\mu$) and standard deviation ($\sigma$), which are subsequently used to generate random coding variables which follow a Gaussian distribution ($G$ in FIG. 3) in the latent space. The coding variables are then fed to the decoder network for trajectory reconstruction.

The decoder network, which comprises three layers of t-CNN, reconstructs the data following the inverse process of the encoding CNNs. The output of the decoder network is the generated trajectory datasets, which will be written as $\hat{s}(\hat{r}, \hat{\theta}, \hat{z}, t)$ in later sections. Again, in the following text, we occasionally use $\hat{s}(t)$ to generically denote the cylindrical components $\hat{r}(t)$, $\hat{\theta}(t)$, and $\hat{z}(t)$ of the generated trajectories.

### C. Objective function and model training

The objective function was designed to guarantee the accuracy and generalization criteria. The accuracy criterion requires that the generated trajectories are statistically representative of that obtained in experiments, and such an objective, denoted as $J_A(\vartheta)$, is implemented using the mean squared Euclidean norm of the distance between the input and output trajectories:

$$J_A(\vartheta) = \frac{t_1}{mt_n}\sum_{i=1}^{m}\sum_{t=t_1}^{t_n}\|s_i(t) - \hat{s}_i(t)\|_2, \qquad (1)$$

where $\vartheta$ denotes all the variables which are to be determined in the model training (not to be confused with the azimuthal coordinate $\theta$). One can observe that the minimization of Eq. (1) leads to the scenario that the generated trajectories are exactly the ground truth plus some degree of variance due to information loss. This trivial output should be avoided, and thus we emphasize the generalization criterion. In other words, the generated trajectories should manifest a sufficient degree of diversity, so that they are regarded as new events independent of the inputs. Such diversity can be rendered in the latent space, wherein the coding variable, in our application, is generated per a simple Gaussian distribution. Mathematically, such a condition requires minimization on the Kullback-Leibler (KL) divergence between the coding distribution and a unity Gaussian function, which, after simplification [17,27], yields the objective $J_G(\vartheta)$:

$$J_G(\vartheta) = \frac{1}{2m}\sum_{i=1}^{m}[\mu^2 + \sigma^2 - 1 - \log(\sigma)]. \qquad (2)$$

Combining Eqs. (1) and (2) provides us the objective function $J(\vartheta)$ for the training of the VAE model:

$$J(\vartheta) = J_A(\vartheta) + wJ_G(\vartheta), \qquad (3)$$

where $w$ is a weight coefficient taking positive values less than unity, and is used to allocate the amount of emphasis placed on accuracy and generalization.

The training of the VAE model is conducted using stochastic gradient descent (the number of minibatches is the size of the overall batch divided by the size of a minibatch), and the gradient is computed using back propagation. In total, fifteen batches of experimentally determined trajectories are used for the model training. The procedure is outlined in Table 1.

Table 1. VAE training algorithm

For **number of iterations,** do
    For **number of minibatches,** do
- **Randomly pick m trajectories $\{s_1, s_2, \cdots, s_m\}$**
- **Use forward pass (from $s(t)$ to $\hat{s}(t)$) to compute $\{\hat{s}_1, \hat{s}_2, \cdots, \hat{s}_m\}$**
- **Compute the objective function from eq. 3**
- **Use back propagation to compute the gradients for model parameters $\nabla_\vartheta J(\vartheta)$**
- **Update the model parameters $\vartheta$**

    End for
End for

## III. RESULTS AND DISCUSSION

Figure 5 (a), (b), and (c) respectively show the trajectories which were generated using the VAE model under conditions of $w = 10^{-3.5}$, $10^{-4}$ and $10^{-5}$. Panel (d) plots the trajectories which were determined in our particle tracking experiments (hereafter, the ground truth). The visual similarity between the simulated trajectories and the ground truth increases with decreasing $w$, because more weight has been placed on the accuracy aspect in the objective function. The model trained under $w = 10^{-3.5}$ produces trajectories which appear to be "under-developed", as none of the $r$, $\theta$, or $z$ components reaches the corresponding length-scale of the ground truth. When $w$ increases to $10^{-5}$, the differences between the VAE generated trajectories and the ground truth are barely distinguishable. Nevertheless, trajectories in (c) and (d) start to share repetitive appearances, indicating a possible overfitting condition in which the generated trajectories depend heavily on the model inputs.

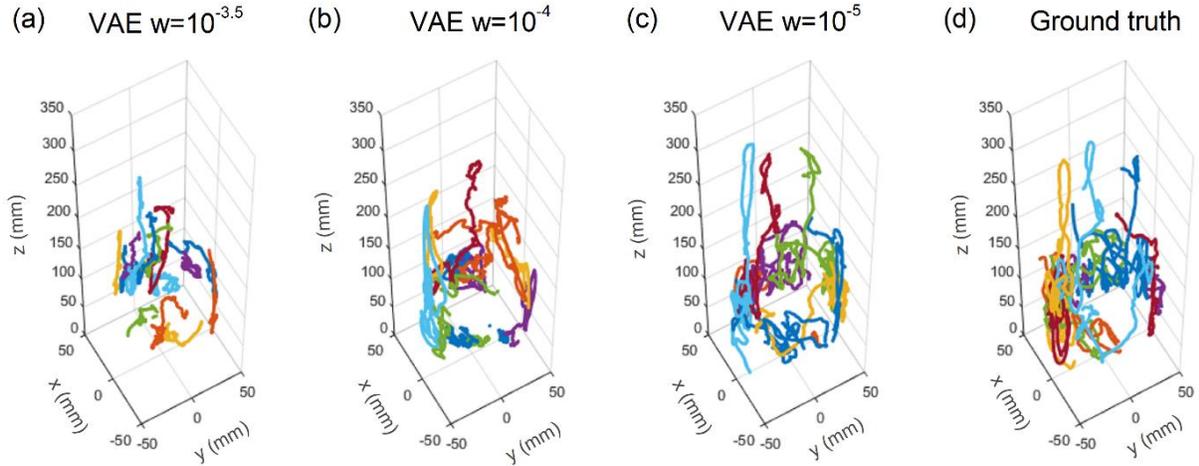

**FIG. 5. The trajectories generated using VAE model are compared with the ground truth. (a), (b) and (c) respectively show the trajectories generated under optimization conditions of $w = 10^{-3.5}$, $10^{-4}$, and $10^{-5}$. (d) shows the trajectories obtained from our tracking experiments.**

We next quantitatively evaluate the accuracy of the generated trajectories by comparing their mean squared displacements ($MSD_{\hat{s}}$) to that of the ground truth ($MSD_s$), which is shown in FIG. 6. The calculation of $MSD_{\hat{s}}$ for the simulated motion in direction $\hat{s}$, during a finite time interval ($\Delta t$), follows the time-averaged manner introduced in Refs [16,28]:

$$MSD_{\hat{s},i}(\Delta t) = \frac{t_0}{t_n - \Delta t} \sum_{t=t_0}^{t_n - \Delta t} [\hat{s}_i(t + \Delta t) - \hat{s}_i(t)]^2, \tag{4.1}$$

$$\text{and, } MSD_{\hat{s}}(\Delta t) = \frac{1}{m} \sum_{i=1}^{m} MSD_{\hat{s},i}(\Delta t), \tag{4.2}$$

where the $\Delta t$ takes only discrete values that are divisible by $t_0$. Note that we also calculate the $MSD_s$ for the ground truth using Eq. (4) with $s(t)$ instead of $\hat{s}(t)$. Figure 6 shows the scaling relationships between $MSD_s$ (or $MSD_{\hat{s}}$) and $\Delta t$ in log-log spaces. (Hereafter, the subscripts of $MSD$ will be omitted, if the same analysis is performed on the VAE generated trajectories and ground truth, or if there is no need to distinguish directions). Power-law relationships are observed in all directions, which can be written as following:

$$MSD \propto \Delta t^{\gamma}, \tag{5}$$

where $\gamma$ is a scaling exponent that indicates whether the motion is ballistic or diffusive, taking values of two or one, respectively [9,16,29]. The motion in all directions starts out ballistically, and inflection starts to occur when the timescale is sufficiently large, indicating the washing-out of correlations. In the $r$ and $z$ directions, the $\gamma$ asymptote to zero near the large $\Delta t$ limit, giving an invariant $MSD$, because the motion of particles is ultimately bounded in a system with a finite length-scale. The motion in the $\theta$ direction is circular and unbounded [30], so that normal diffusion manifests ($\gamma = 1$) when the motion is fully randomized at large $\Delta t$. By comparing the generated trajectories and the ground truth, we can see that the inflections in the power-law, and subsequently the long-term behaviors – represented by $\gamma$ – of the motion, are successfully reproduced by the VAE model. The "under-developed" nature of the generated trajectories is again observed with larger $w$ values which produces trajectories reaching smaller $MSD$. Similar behavior can be observed in the velocity distributions $p(v_{\hat{s}})$ calculated from the model outputs, which are also compared with the ground truth $p(v_s)$ in FIG. 7. As $w$ increases from $10^{-5}$ to $10^{-3.5}$, the deviation between the VAE generated trajectories and the ground truth becomes pronounced.

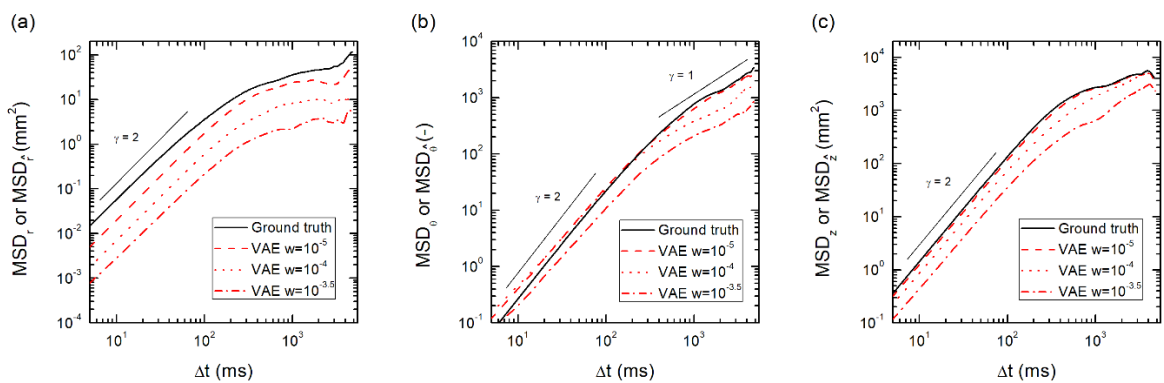

**FIG. 6.** The scaling relationships between $MSD$ and $\Delta t$ in the (a) $r$, (b) $\theta$, and (c) $z$ directions. Black solid lines represent the $MSD_s$ calculated from the ground truth. Red dashed lines, dotted lines, and dashed-dotted lines respectively represent the $MSD_{\hat{s}}$ calculated from VAE generated trajectories under $w = 10^{-3.5}, 10^{-4},$ and $10^{-5}$.

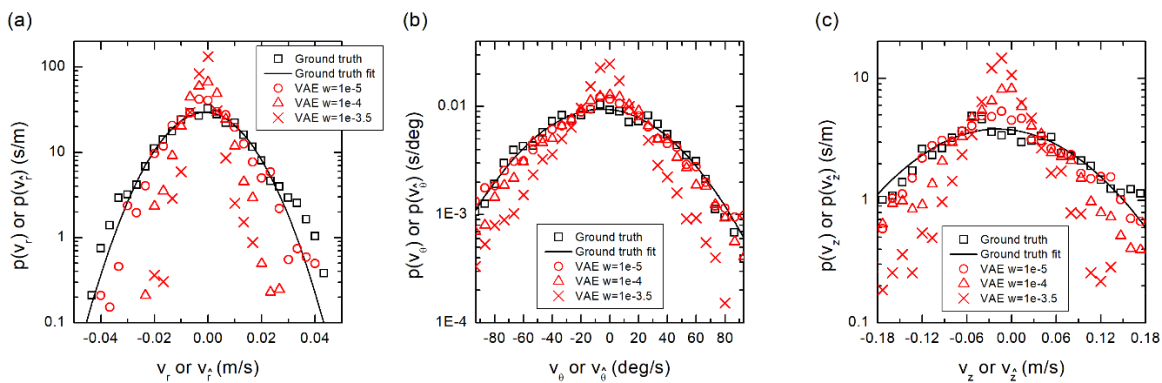

**FIG. 7** Velocity distribution in (a) $r$, (b) $\theta$, and (c) $z$ directions. Squares represent the probability density distribution of the velocities calculated from the ground truth $p(v_s)$, and black lines represent a least square fitting per normal distribution function. Circle, triangle, and cross symbols respectively represent the velocities calculated from the VAE generated $p(v_{\hat{s}})$ under $w = 10^{-3.5}, 10^{-4},$ and $10^{-5}$.

We next quantify the generalization of VAE outputs by calculating the Pearson correlation coefficients $\zeta(s_i, \hat{s}_j)$ between the generated trajectories and the ground truth:

$$\zeta(s_i, \hat{s}_j) = \frac{\mathbb{E}[(s_i - \mathbb{E}[s_i])(\hat{s}_j - \mathbb{E}[\hat{s}_j])]}{\mathbb{E}[(s_i - \mathbb{E}[s_i])^2]^{1/2} \mathbb{E}\left[(\hat{s}_j - \mathbb{E}[\hat{s}_j])^2\right]^{1/2}}, \quad (5)$$

where $\mathbb{E}[,]$ denotes calculating the expectation values of the variables in the bracket [17]. FIG. 8 maps the $\zeta(s_i, \hat{s}_j)$ matrices calculated between the ground truth and VAE generated trajectories under $w = 10^{-5}$ (panels d-f), $10^{-4}$ (panels g-i), and $10^{-3.5}$ (panels j-l). For each case, the correlations in the $r$, $\theta$ and $z$ spaces are individually calculated and presented in each sub-panel. The self-correlations of ground truth, serving as a baseline for the discussion of generalization, were also calculated. These self-correlation matrices, written as $\zeta(s_i, s_j)$, are presented in FIG. 8 (a)-(c).

We observe that (a)-(c) are symmetric according to the principal diagonal lines ($i = j$), on which the elements take a value of unity. The principal diagonal lines represent cases in which two events of interest are identical. The rest of the elements, which are not on the diagonal ($i \neq j$, hereafter, the *background*) provide us the degree of correlation between the independent events determined in experiments. FIG. (d)-(f) show that when $w$ is as low as $10^{-5}$, each generated trajectory (in all directions) can always find a highly-correlated counterpart in the ground truth, (see the scattered red-colored "*hot spots*"). This degree of correlation suggests the generated trajectories are mostly not new events. When $w$ increases to $10^{-4}$, significant improvement can be observed (FIG.9 (g)-(i)) as most of the *hot spots* disappear. When $w$ is as large as $10^{-3.5}$ (FIG. 9 (j)-(l)), the values of the correlation coefficients approach the *background* level in (a)-(c).

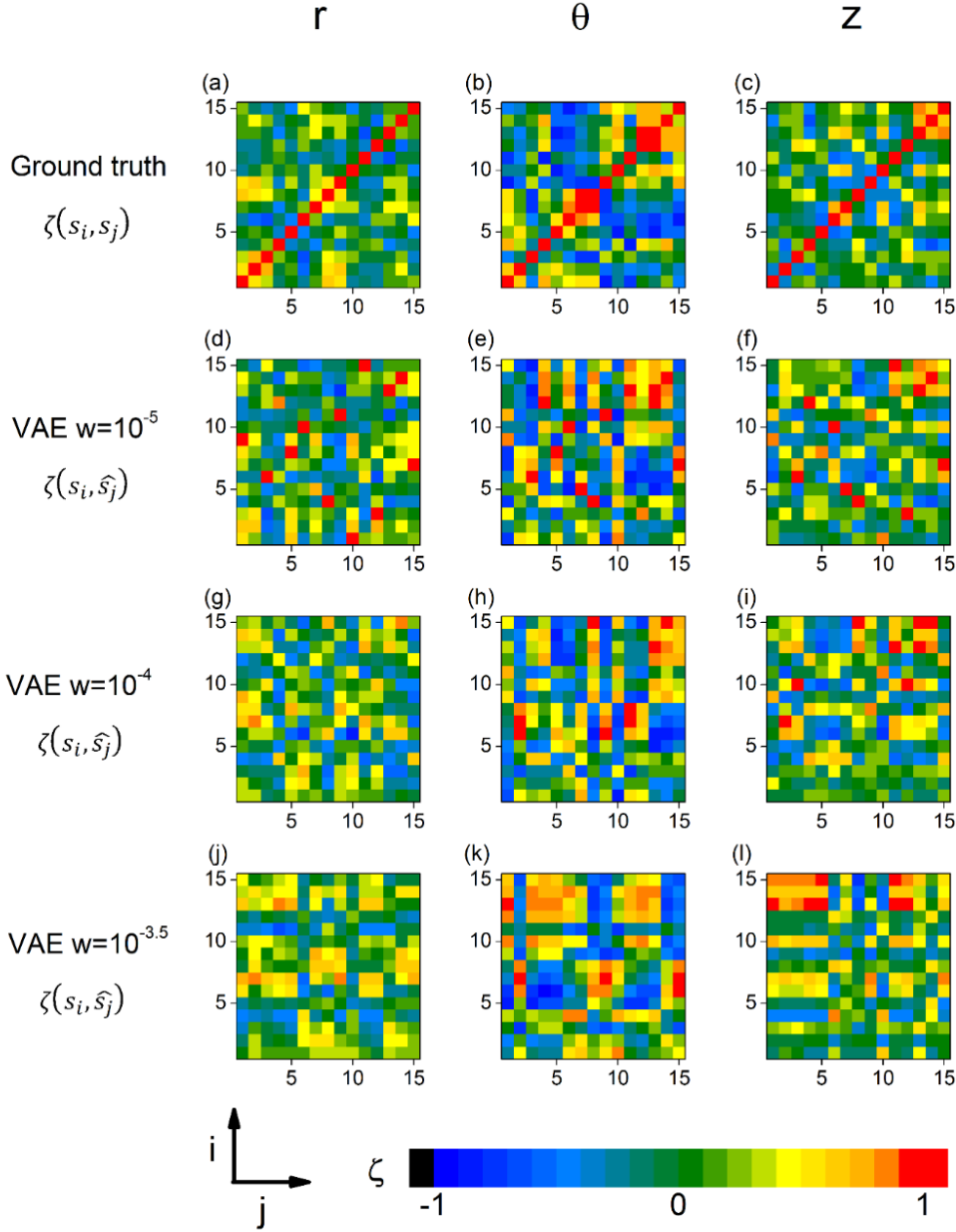

**FIG. 8. Two-dimensional color maps for the Pearson correlation coefficient matrices. (a), (b) and (c) respectively show $\zeta(r_i, r_j)$, $\zeta(\theta_i, \theta_j)$, and $\zeta(z_i, z_j)$, which are the self-correlations among the 15 batches of experimentally determined trajectory datasets. (d), (e) and (f) respectively show $\zeta(r_i, \hat{r}_j)$, $\zeta(\theta_i, \hat{\theta}_j)$ and $\zeta(z_i, \hat{z}_j)$, which are the correlations between the experimentally determined trejactories and the VAE generated trajectories under $w = 10^{-5}$. (g) - (i) show the correlation between the experimentally determined trejactories and VAE generated under $w = 10^{-4}$, and (j) - (l) show the correlation matrices for the case of $w = 10^{-3.5}$. In panels (d) - (l), the horizontal axis represents the identification of the VAE generated trajectories, denoted by $j$, and the vertical axis represents the identification of the experimentally obtained trajectories, denoted by $i$.**

We now stich together our main findings in FIG. 9. The VAE outputs with $w$ varying across five orders of magnitude are scored in a *Accuracy – Generalization* space. The *Accuracy* factor is calculated using the $MSD(\Delta t)$ statistics:

$$Accuracy = 1 - \langle \eta_s \rangle, \tag{6.1}$$

$$\text{with } \langle \eta_s \rangle = \frac{t_1}{t_n - t_1} \sum_{\Delta t = t_1}^{t_n - t_1} \left| \frac{\log_{10} MSD_s(\Delta t) - \log_{10} MSD_{\hat{s}}(\Delta t)}{\log_{10} MSD_s(\Delta t)} \right|. \tag{6.2}$$

When calcualting $\langle \eta \rangle$ in 3-dimensional space, the $MSD_s(\Delta t)$ and $MSD_{\hat{s}}(\Delta t)$ in (6.2) are replaced by their 3-dimensional counterparts.

The *Generalization* factor is calculated based on the Pearson correlation coefficient matrices:

$$Generalization = 1 - \langle \zeta_s \rangle, \tag{7.1}$$

$$\text{with } \langle \zeta_s \rangle = \frac{1}{m} \sum_{j=1}^{m} \left[ \max_i \zeta_s(s_i, \hat{s}_j) \right]. \tag{7.2}$$

When calculating $\langle \zeta \rangle$ in 3-dimensional space, $\zeta_s$ in (7.2) is replaced by the arithmatic mean matrix $\frac{1}{3}(\zeta_r + \zeta_\theta + \zeta_z)$.

Figure 9 shows the tradeoff between accuracy and generalization. Tthe model performs best with $w$ = ca. $10^{-4}$. Theoretically, the performance of the current VAE model could be improved by fine tuning the neural network structure and/or the training procedures [17,23,27], for example by constructing separate neural networks for each cylindrical coordinate, using a deeper model structure, or applying different activation functions. However, here we also need to emphasize the fundamental limitations at the core of the VAE model. First, the VAE model determines similarity based on quantified factors – in this case, Euclidean distance between the input and output datasets – a stiff mechanism which often leads to overfittings. In other words, the generated trajectories are *exact* instead of being *similar*. Such a mechanism is not compatible with the objective of generalization, and compromises between accuarcy and generalization are required, which eventually limit the model's applicability. Second, trajectories generated using the VAE model is unconstrained on their initial coordinates, and as a result, propagation of existing trajectories is not controllable. Third, the encoder-decoder architecture requires indentical dimensionality for the input and output datasets, and thus the lengths of generated trajectories can not exceed those of the input datasets.

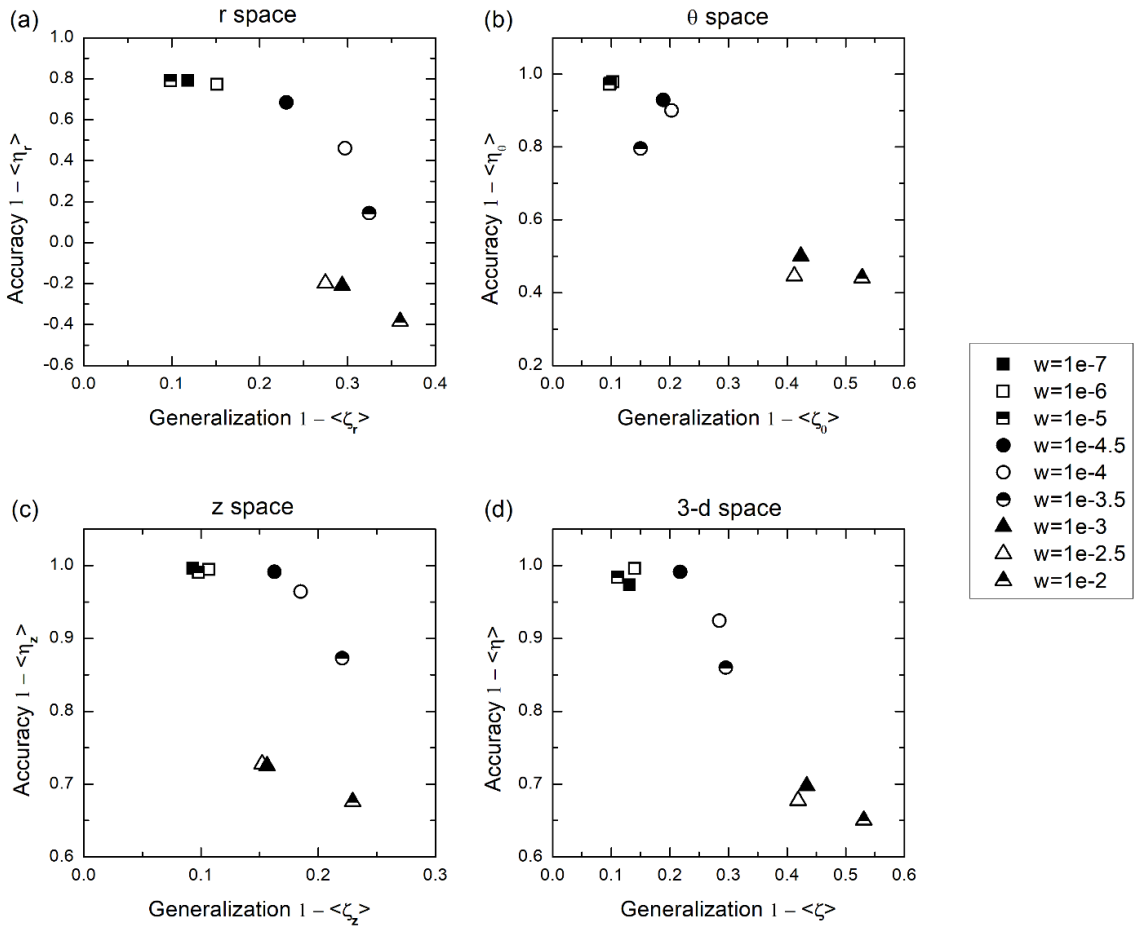

**FIG. 9. Evaluating the performance of the VAE model with respect to accuracy and generalization. (a), (b), and (c) respectively show the results calculated using the r, θ, and z components of the trajectories. (d) shows the result calculated using the trajectories in 3-dimensional space. Data symbols of various shape represent changing optimization conditions, with w varying across five decades.**

# CONCLUSION

In this work, we introduce a deep learning model, based on variational autoencoder neural networks, for simulating of Lagrangian motions of stochastic particles trapped in a flame system. Our results show that the model succeeds in generating trajectories which are statistically representative to that determined in experiments. The hyper-parameter space was carefully explored toward reaching a configuration which balances the accuracy and generalization of the model outputs. Limitations associated with VAE model, which commonly serves as a starting point of designing a generative model, are discussed. In future research, attempts will be made to resolve these limitations and improve on the generalization of the model output. We also emphasize that the trajectory simulation discussed in this work is statistically based and data driven. Future research should be directed toward incorporating physical models in the deep learning network, so that the characteristics which are extracted from experiments can be understood more fundamentally.

# ACKNOWLEDGEMENTS

The research is funded by US National Science Foundation (NSF) grants AGS-1455215 and CBET-1511964, and the NASA Radiation Science Program grant NNX15AI66G. We thank Benjamin J. Sumlin, Drs. William R. Heinson, Kuan-Yu Shen and Akshay Gopan for many helpful discussions on the design and implementation of the particle tracking apparatus.

# REFERENCES

[1] B. R. Munson *et al., Fundamentals of fluid mechanics* (John Wiley & Sons, New Jersey 2006).
[2] A. R. Rammohan *et al.,* Chem. Eng. Sci. **56,** 2629-2639 (2001).
[3] H. Aref, Phil. Trans. R. Soc. Lond. A **333,** 273-288 (1990).
[4] H. Aref, Phys. Fluids **14,** 1315-1325 (2002).

[5] P. Meakin *et al.,* Phys. Rev. A **34,** 5091 (1986).
[6] P. Meakin, Phys. Rev. Lett. **51,** 1119 (1983).
[7] W. R. Heinson, P. Liu, and R. K. Chakrabarty, Aerosol Sci. Technol. **51,** 12-19 (2017).
[8] W. R. Heinson *et al.,* Aerosol Sci. Technol. **52,** 953-956 (2018).
[9] F. Bartumeus *et al.,* Ecology **86,** 3078-3087 (2005).
[10] D. W. Sims *et al.,* Nature **451,** 1098 (2008).
[11] N. E. Humphries *et al.,* Nature **465,** 1066 (2010).
[12] B. Jiang and T. Jia, Int. J. Geogr. Inf. Sci. **25,** 51-64. (2011).
[13] P. M. Torrens *et al.,* Comput. Environ. Urban **36**, 1-17 (2012).
[14] D. S. Ullman *et al.,* J. Geophys. Res. Oceans **111,** C12 (2006).
[15] J. D. Paduan and P. P. Niiler, J. Geophys. Res. Oceans **95,** C10 (1990).
[16] P. Liu *et al.,* Physical Review E **97,** 042102 (2018).
[17] I. Goodfellow, Y. Bengio, and A. Courville, *Deep learning* (Massachusetts Institute of Technology Press, Massachusetts 2016).
[18] C. X. Hernández *et al.,* arXiv: 1711.08576 (2017).
[19] R. Sharma *et al.,* arXiv:1807.11374 (2018).
[20] S. J. Wetzel, Phys. Rev. E **96,** 022140 (2017).
[21] J. N. Kutz, J. Fluid Mech. **814,** 1-4. (2017).
[22] I. Goodfellow, arXiv:1701.00160 (2016).
[23] C. Doersch, arXiv:1606.05908 (2016).
[24] R. K. Chakrabarty *et al.,* Appl. Phys. Lett. **104,** 243103 (2014).
[25] P. Liu *et al.,* Aerosol Sci. Technol. **49,** 2732-1241 (2015).
[26] P. Liu, W. R. Heinson, and R. K. Chakrabarty, Aerosol Sci. Technol. **51,** 879-886 (2017).
[27] D. P. Kingma and M. Welling, arXiv:1312.6114 (2013).
[28] R. Metzler *et al.,* Phys. Chem. Chem. Phys. **16,** 24128-24164 (2014).
[29] T. H. Solomon, E. R. Weeks, and H. L. Swinney, Phys. Rev. Lett. **71,** 3975 (1993).
[30] A. Lee, Wiley Interdiscip. Rev. Comput. Stat. **2,** 477-486 (2010).